\documentclass[11pt,a4paper]{article}
\usepackage[hyperref]{ranlp2021}
\usepackage{times}
\usepackage{latexsym}

\usepackage{graphicx}
\usepackage{graphics}
\usepackage{subcaption}
\usepackage{amsmath}
\usepackage{multirow}
\usepackage{multicol}
\usepackage{enumitem}
\usepackage{makecell}
\usepackage{bm}
\usepackage[belowskip=-5pt,aboveskip=2pt,font=10pt]{caption}
\aclfinalcopy 
\def\aclpaperid{100} 

\title{GeSERA: General-domain Summary Evaluation by Relevance Analysis}

\author{Jessica L\'opez Espejel$^{\ast\dagger}$, Gaël de Chalendar$^{\ast}$,  Jorge Garcia Flores$^{\dagger}$,  \\\textbf{Thierry Charnois}$^{\dagger}$, \textbf{and Ivan Vladimir Meza Ruiz}$^{\ddagger}$ \\

    $^{\ast}$ Université Paris-Saclay, CEA, List, F-91120, Palaiseau, France\\
    $^{\dagger}$ CNRS-LIPN-Université Sorbonne Paris Nord, France\\
    $^{\ddagger}$ IIMAS, UNAM, Mexico\\

    jessicanayeli.lopezespejel@univ-paris13.fr, gael.de-chalendar@cea.fr,\\
    jgflores@lipn.fr, thierry.charnois@lipn.univ-paris13.fr, ivanvladimir@turing.iimas.unam.mx\\

  \\}

\date{}

\begin{document}
\maketitle
\begin{abstract}

We present GeSERA, an open-source improved version of SERA for evaluating automatic extractive and abstractive summaries from the general domain. SERA is based on a search engine that compares candidate and reference summaries (called queries) against an information retrieval document base (called index). SERA was originally designed for the biomedical domain only, where it showed a better correlation with manual methods than the widely used lexical-based ROUGE method. In this paper, we take out SERA from the biomedical domain to the general one by adapting its content-based method to successfully evaluate summaries from the general domain. First, we improve the query reformulation strategy with POS Tags analysis of general-domain corpora. Second, we replace the biomedical index used in SERA with two article collections from AQUAINT-2 and Wikipedia. 
We conduct experiments with TAC2008, TAC2009, and CNNDM datasets. Results show that, in most cases, GeSERA achieves higher correlations with manual evaluation methods than SERA, while it reduces its gap with  ROUGE for general-domain summary evaluation. GeSERA even surpasses ROUGE in two cases of TAC2009. Finally, we conduct extensive experiments and provide a comprehensive study of the impact of human annotators and the index size on summary evaluation with SERA and GeSERA.

\end{abstract}

\section{Introduction}

    
    Automatic summary evaluation is a challenging task in Natural Language Processing (NLP). Evaluation is usually done by humans, but manual evaluation is subjective, costly and time expensive~\cite{lin-hovy-2002-manual}. 
    Automatic evaluation methods~\cite{Lin04rougea, Torres2010_FRESA, zhao2019_moverScore,  Zhang2020_BERTScore} are an alternative to save time for users who extract the most relevant content from the web using Automatic Text Summarization systems (ATS). 
   There exist two types of evaluation approaches: (1) manual evaluation methods like Pyramid~\cite{NenkovaP04} and Responsiveness, where human intervention is mandatory, and (2) automatic evaluation methods, where human intervention can be needed as a ground-truth reference~\cite{Lin04rougea, Cohan2016} or not~\cite{Torres2010_FRESA, CabreraDiego2018_SummTriver}.
   
   \textit{Summary Evaluation by Relevance Analysis} (SERA)~\cite{Cohan2016} is an automatic evaluation method that partially relies on human references to evaluate 
   abstractive summaries from the biomedical domain. It was proposed as an alternative to ROUGE~\citep{Lin04rougea}, the widely used automatic metric, that is based on lexical overlaps between candidate and reference summaries. ROUGE is unfair to evaluate abstractive summaries where the ATS paraphrases the text instead of just copying-pasting chunks of it~\cite{Cohan2016}.
   
   SERA is based on content relevance and is fairer to evaluate abstractive summaries because it attributes high scores to summaries that are lexically different but semantically related. However, it surpasses ROUGE on the biomedical domain only. In this paper, we modify SERA to make it usable for generic collections. We propose the following contributions:

        \noindent \textbf{1.} Implement an open-source version of SERA from scratch.
        
        \noindent \textbf{2.}
        Propose GeSERA (\textit{General-domain SERA}), an improved version of SERA that is domain-independent.
        
        \noindent \textbf{3.}  Conduct extensive experiments with two large indexes (AQUAINT-2~\cite{graff2002aquaint} and Wikipedia) and three summarization datasets (TAC\footnote{\url{https://tac.nist.gov/}} 2008, TAC2009, CNN-Daily Mail~\cite{Bhandari2020_re-evaluating}). These datasets are well-suited for general-domain and news abstractive summary evaluation. GeSERA achieves competitive results compared to a range of state-of-the-art evaluation approaches on both abstractive and extractive summaries.

        \noindent \textbf{4.}  Make the code and our Wikipedia dataset publicly available to help future researchers.\footnote{\tiny \url{https://github.com/JessicaLopezEspejel/GeSERA/}} 

        \noindent \textbf{5.}  Conduct extensive experiments on the impact of the index size and human annotators on summary evaluation with SERA and GeSERA.
        

\section{Proposed approach}
\label{sec:proposed_approach}
    
    \vspace{-0.1em}
    \subsection{Baseline: SERA}
    
    \vspace{-0.1em}
    SERA~\cite{Cohan2016} is based on a content relevance between a
    candidate summary and its corresponding human-written reference summaries using information retrieval. SERA compares these summaries (called queries) against a set of documents from the same domain (called index), and compares the overlap of retrieved results.  SERA refines queries in three different manners (1) \textit{Raw text} - only stop words and numbers are removed, (2) \textit{Noun phrases (NP)} - only noun phrases are kept, and  (3) \textit{Keywords (KW)} - only unigrams, bigrams, and trigrams are kept. 
    
    SERA is defined in Equation~\ref{eq:SERA}.
    

    \begin{equation}
        \small
        SERA = \frac{1}{M} \sum_{i=1}^{M}\frac{|R_{C} \cap R_{G_{i}}|}{|R_{C}|}
    \label{eq:SERA}
    \end{equation}
    

    where: $R_{C}$ is the list of retrieved documents for the candidate summary $C$, $R_{G_{i}}$ is the ranked list of retrieved documents for the reference summary $G_i$, and $M$ is the number of reference summaries.
    
    Another variant of SERA is called SERA-DIS. It takes the order of retrieved documents into consideration (Equation~\ref{eq:SERA_DIS}).
    
    \begin{equation}
    \small
    \label{eq:SERA_DIS}
      \begin{aligned}
        SERA-DIS= \frac{ \sum_{i=1}^{M}  (\sum_{j=1}^{|R_{C}|} \sum_{k=1}^{|R_{G_{i}}|}~X_{j, k}}{M * D_{max}} \\
        \\
        X_{j, k}=  \left\lbrace
                \begin{array}{ll}
                \frac{1}{log(|j-k|+2)} & if~ R_{C}^{(j)} = R_{G_{i}}^{(k)} \\
                 0 & otherwise 
                \end{array}
                \right.
     \end{aligned}
    \end{equation}
    
    where: $R^{(j)}_C$ is the $j^{th}$ result in the ranked list $R_C$, and $D_{max}$ is the maximum achievable score used for normalization. In both SERA variants, retrieved results are truncated at 5 and 10 documents (hence the notations SERA-5 and SERA-10 in Section~\ref{sec:experiments}). ~\citet{Cohan2016} used articles from PubMed\footnote{\url{http:// www.ncbi.nlm.nih.gov/pmc/}} as a index, and summaries from TAC 2014 as queries.

    
    The intuition behind SERA is that a summary context is represented by its most related articles. Thus, two summaries related to the same documents are semantically related, even if they are lexically different. Consequently, SERA is fairer to evaluate abstractive summaries contrarily to the lexical-based ROUGE. However, SERA suffers from a series of limitations: (1) the code is not open-source, (2) no information was provided concerning the subset of PubMed used as an index, and (3) PubMed is specialized in the biomedical domain only. The first two drawbacks make SERA unusable by the community, while the third restricts its usage to the biomedical domain.
    

    \subsection{GeSERA: General-domain SERA}
    \label{subsec:wiki_sera}
    
    We build on SERA merits and limitations to propose GeSERA, an open-source version of SERA that evaluates summaries from the general domain. Novelties of GeSERA are the index pool and query reformulation adapted to the evaluation of summaries from the general domain.

    \paragraph*{Index documents pool -}
    Differently from SERA, GeSERA enables general-domain summary evaluation. It is thus necessary to replace the biomedical index used by~\citet{Cohan2016} with a set of documents related to the general domain. We build many indexes using a variant number of articles from Wikipedia and AQUAINT-2. See Subsection~\ref{subsec:index_datasets} for more details.

    \paragraph*{Query reformulation (QR) -} It
    improves retrieval process by removing unnecessary terms from the text. Therefore, we propose a different approach to refine queries in GeSERA that is better suited for general domain summaries. 
    
\begin{figure}[htb!]
        \centering
        
        \begin{minipage}{.38\textwidth}
          \centering 
          \captionsetup{width=.9\linewidth}
          \includegraphics[width=.999\linewidth]{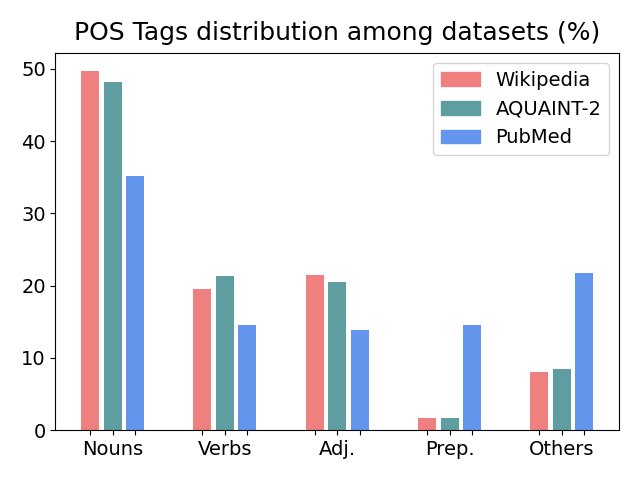}
          
        \end{minipage}%
    
    \caption{POS Tags distribution percentages for Wikipedia, AQUAINT-2, and PubMed datasets}
    \label{fig:datasets_statistics}
    \end{figure}

    According to~\citet{Tan2020_summarizationCOVID}, nouns in generated summaries represent more accurately the information conveyed by the original abstracts than other POS tags. This study was conducted on Covid-19 medical texts and can explain why SERA achieved a higher correlation than ROUGE for the TAC 2014 biomedical dataset. We led an analysis that consists of analyzing Part-Of-Speech (POS) tags distribution for PubMed (biomedical dataset built by~\citet{cohan2018_pubmed}), AQUAINT-2 (news corpus), and Wikipedia (general domain encyclopedia). Figure~\ref{fig:datasets_statistics} shows bar plots for percentages of nouns, verbs, adjectives (Adj.), prepositions (Prep.), and the total percentage of other tags (Others).
    
    Our analysis of three datasets which describe different domains confirms the observation of~\citet{Tan2020_summarizationCOVID} for PubMed. However, it shows that the percentages of verbs and adjectives are higher in AQUAINT-2 and Wikipedia than in PubMed. Equally important, there is a remarkable absence of prepositions in Wikipedia and AQUAINT-2. Based on our analysis, we propose to reformulate queries in GeSERA by only keeping tokens tagged with  nouns, verbs, and adjectives, the three most frequent tags in the news and  general domain corpora.
    
    \paragraph*{Search engine -} A semantic-based retrieval approach is crucial when handling abstractive summaries.
    In order to compare the queries against the index, a search engine is needed for information retrieval and scoring. We use the Whoosh\footnote{\url{https://whoosh.readthedocs.io/en/latest/intro.html}} search engine with the BM25F (\textit{Best Match 25 Model with Extension to Multiple Weighted Fields}) ranking function~\citep{DBLP:conf/trec/ZaragozaCTSR04}. This model is widely used for semantic search~\cite{perez2010using, DBLP:journals/ftir/RobertsonZ09}. It consists in weighting terms according to their field importance, combining them, and using the resulting pseudo-frequencies for ranking. 
    
    
\section{Experiments}
\label{sec:experiments}

    SERA was developed in the context of scientific biomedical article summarization with the idea that its semantic specificity is particularly useful for this domain. We hypothesize that if we reformulate queries properly and change the index pool, SERA can assess summaries from other domains for both abstractive and extractive summarization. This hypothesis is based on the fact that SERA considers terms that are not lexically equivalent but are semantically related. We conduct extensive experiments on SERA and GeSERA to test our hypothesis.

    \subsection{Index datasets}
    \label{subsec:index_datasets}
    
    The index is a key component of GeSERA approach insofar as it should describe the same domain as the queries. The number of documents in the index is also decisive as we will show in Subsection~\ref{subsec:impact_index}. We describe briefly query and index datasets hereafter and provide more information in the supplementary material. 
    
    \begin{itemize}[leftmargin=*]
    \setlength\itemsep{-0.3em}
    
    \item \textbf{AQUAINT-2}~\cite{graff2002aquaint} is a news corpus built from various sources. We vary the size of the index to include $\mathcal{I}=$\{$10000$, $15000$, $30000$, $60000$, $89760$, $179520$, $825148$\} randomly-selected documents.

    \item \textbf{Wikipedia} is a free encyclopedia that contains various information from the general domain. We vary the size of the index to include $\mathcal{I}=$\{$10000$, $15000$, $30000$, $1778742$\} randomly-selected documents.
    
    \end{itemize}
    
    \subsection{Query datasets}
    \label{subsec:query_dataset}
    Candidate and reference summaries from the news datasets TAC2008 and TAC2009, and the CNN Daily Mail (CNNDM) version published by~\citet{Bhandari2020_re-evaluating} are used as queries. 
    
    \textbf{TAC2008} (/and \textbf{TAC2009}) are subsets of AQUAINT-2. They contain 5568 (/4840) candidate summaries proposed by 58 (/55) participants, and 384 (/352) reference summaries, respectively. These datasets are designed for multi-document extractive summarization (one summary is shared by a set of documents).
    
    
    
    \textbf{CNN Daily Mail}~\cite{Bhandari2020_re-evaluating} is a news dataset, and is of great interest to us because it contains candidate summaries obtained from both extractive and abstractive systems. It consists of 100 reference summaries, having each 25 candidate summaries generated by 11 extractive and 14 abstractive systems. It is designed for mono-document abstractive and extractive summarization (one summary for each document). 
    
    \subsection{Baselines}
    \label{subsec:baselines} 
    We compare GeSERA with some of the most influential evaluation metrics from the literature: 
    \begin{itemize}[leftmargin=*]
        \setlength\itemsep{-0.3em}
        \item  \textbf{ROUGE} and  \textbf{SERA} - two automatic evaluation approaches that rely on human intervention. ROUGE has many variants, but we only report the most popular ones: ROUGE-N ($N= \{1, 2\}$ is the n-gram size) and ROUGE-L (Longest Common Subsequence). For each variant, we report the F-score, Recall and Precision. 
        
        \item \textbf{MoverScore}~\cite{zhao2019_moverScore} and \textbf{BERTScore}~\cite{Zhang2020_BERTScore} - two automatic evaluation approaches based on BERT.
        
        \item \textbf{JS-2}~\cite{lin2006_information} - Jensen-Shannon divergence between bigram's distribution of the candidate and reference summaries.
        
        \item \textbf{SummTriver} (ST)~\cite{CabreraDiego2018_SummTriver} - based on trivergence, i.e. a composition ($\mathcal{T}_c$) or multiplication ($\mathcal{T}_m$) of  Kullback-Leibler (KL), Jensen-Shannon (JS), and smoothed Jensen-Shannon (sJS) divergences. It does need a human reference.
        
        \item \textbf{FRESA}~\cite{Torres2010_FRESA} - an automatic evaluation method that does not rely on human intervention. It is based on Jensen-Shannon divergence and has four variants: unigrams (FRESA-1), bigrams (FRESA-2), trigrams (FRESA-3), and skip-grams (FRESA-4). 
    \end{itemize}
    
    \vspace{-0.35em}
    
    More information is in Section~\ref{sec:related_work}.
    Implementation details are in the supplementary material.

    \subsection{Evaluation methodology}
    \label{subsec:evaluation_metric}
    
    To compare GeSERA with other state-of-the-art methods, we measure the correlation between the scores provided by each automatic and manual evaluation methods. The manual evaluation approaches used here are:
    
    \begin{itemize}[leftmargin=*, noitemsep,topsep=0pt]

     \item  \textbf{Pyramid}~\cite{NenkovaP04} exploits the content distribution in human summaries using \textit{Summary Content Units} (SCUs) based on their frequency in the summary corpus.
     
     \item \textbf{LitePyramid} -~\citep{shapira2019_LitePyramid} is a crowdsource-based lightweight version of Pyramid that relies on statistical sampling instead of exhaustive SCU extraction and testing. 
     \item \textbf{Responsiveness} measures the summary's linguistic quality. 
      
     \end{itemize}
      Following~\citet{Cohan2016}, we use the correlation metrics: (1) \textit{Pearson}~\citep{benesty2009_pearson}, (2) \textit{Spearman}~\citep{kokoska2000_spearman}, and (3) \textit{Kendall tau-b}~\citep{kendall1945_treatment}.

        \begin{figure*}[htb!] 
  \begin{minipage}{0.5\linewidth}
    \centering
    \includegraphics[width=\linewidth]{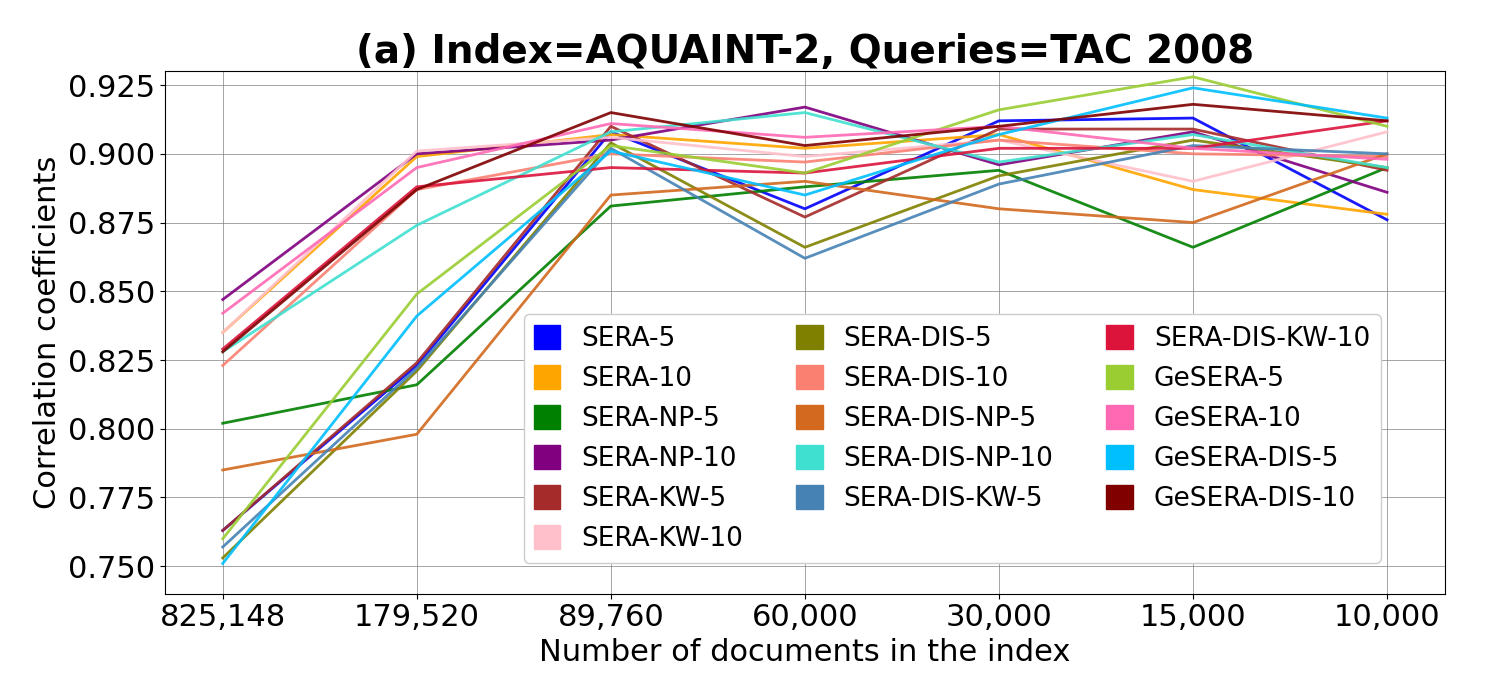} 
  \end{minipage}\hfill
  \begin{minipage}{0.5\linewidth}
    \centering
    \includegraphics[width=\linewidth]{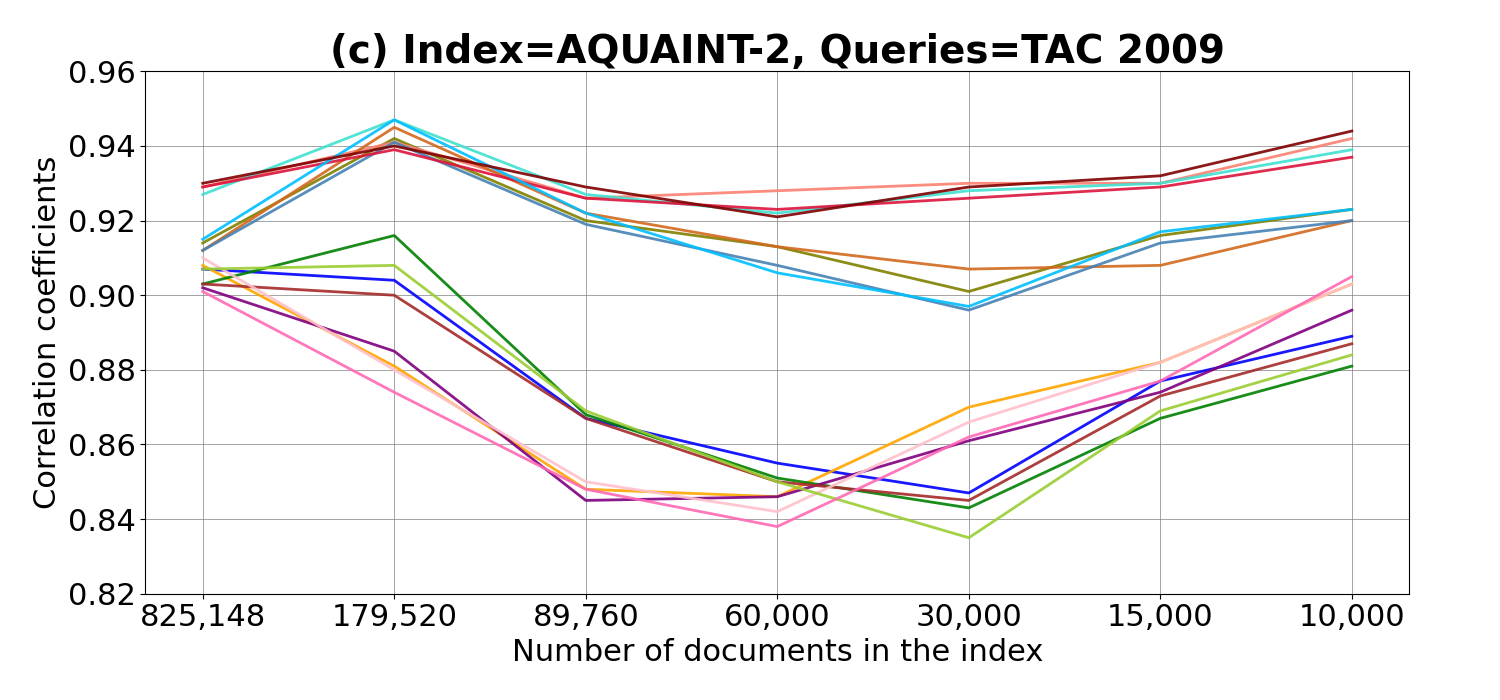} 
  \end{minipage}\hfill
    \begin{minipage}{0.5\linewidth}
    \centering
    \includegraphics[width=\linewidth]{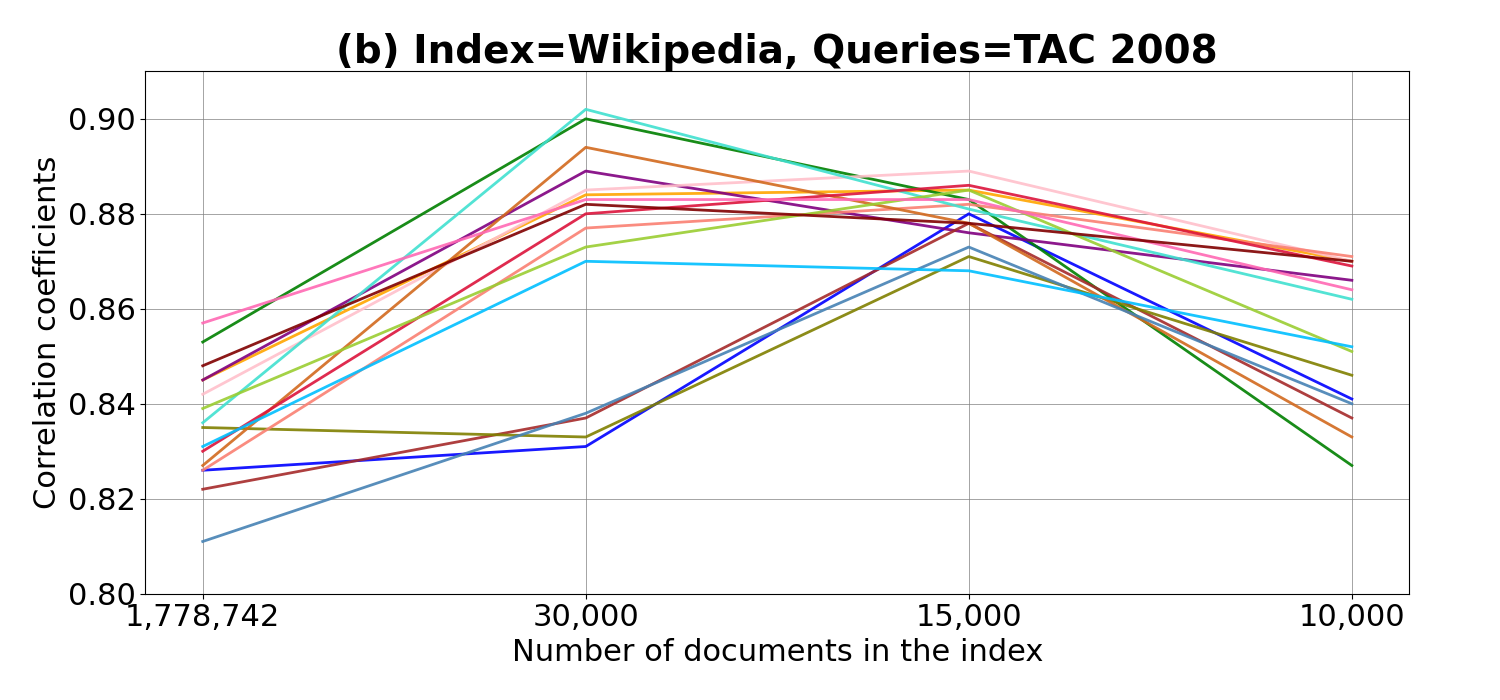} 
  \end{minipage} \hfill
  \begin{minipage}{0.5\linewidth}
    \centering
    \includegraphics[width=\linewidth]{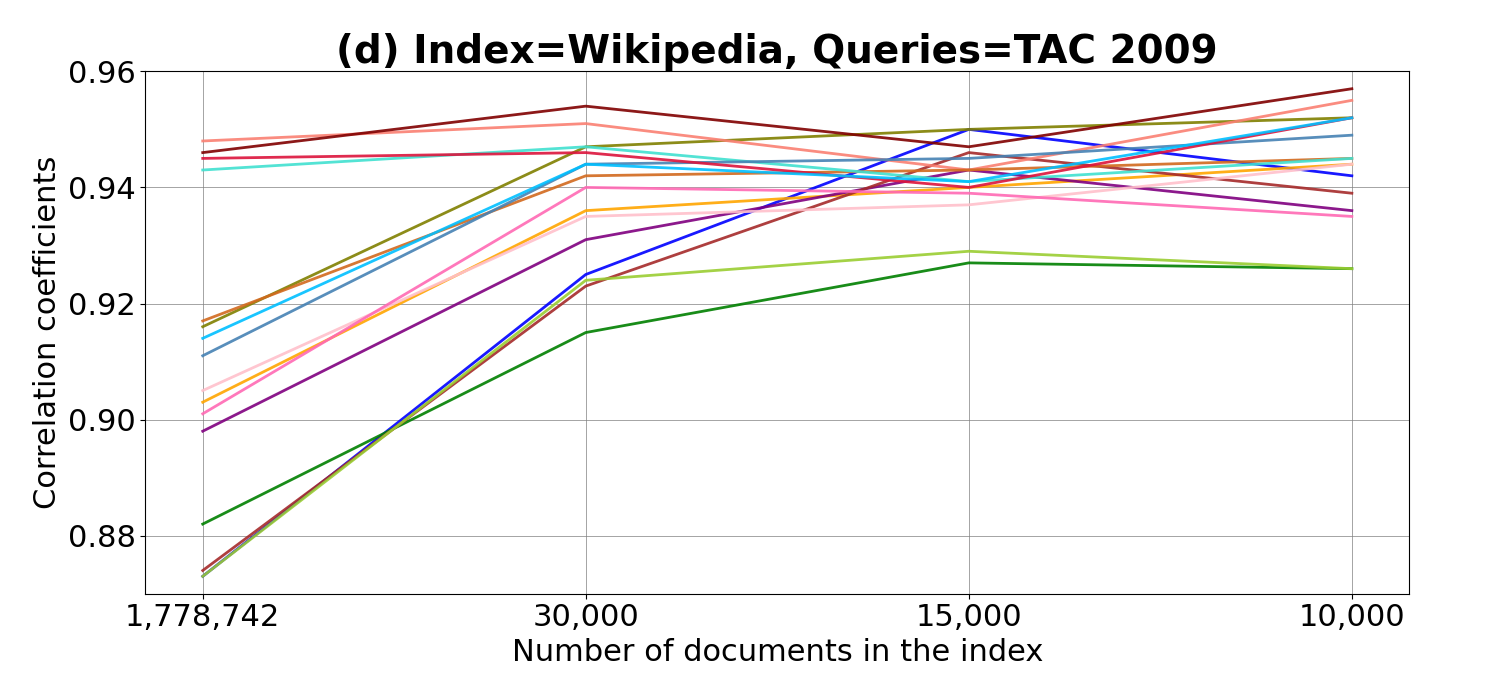} 
  \end{minipage}\hfill
  \caption{Pearson correlation coefficients using TAC2008 and TAC2009 as queries, and AQUAINT-2 and Wikipedia as indexes. Scores are averaged over all human annotators $\mathcal{A}_1, \mathcal{A}_2, \mathcal{A}_3$, and $\mathcal{A}_4$. \textit{Best viewed in color.}}
  \label{fig:pearson_pyramid_aquaint_wikipedia} 
\end{figure*}

\section{Results and discussion}
\label{sec:results}

    In Subsection~\ref{subsec:impact_index}, we vary the index size in SERA and GeSERA and study the variation of their performance on TAC datasets by averaging the score of all four manual annotators $\mathcal{A}_1, \mathcal{A}_2, \mathcal{A}_3$, and $\mathcal{A}_4$. Once we determine the best index size, we present in Subsection~\ref{subsec:main_results} the correlations using the best index size of each method.
    

    
    \vspace{-0.5em}
    
    \subsection{Impact of the index size on the performance of SERA and GeSERA}
    \label{subsec:impact_index}
    
    We first present the impact of the indexes sizes built from Wikipedia on SERA and GeSERA, then the ones built from AQUAINT-2.
   
     \paragraph*{Wikipedia Index - } Figures~\ref{fig:pearson_pyramid_aquaint_wikipedia}-b and~\ref{fig:pearson_pyramid_aquaint_wikipedia}-d show Pearson correlations of SERA and GeSERA with Pyramid when indexing different values of $\mathcal{I}$ from the Wikipedia dataset, and when using TAC2008 and TAC2009 as query datasets, respectively. Figures show that the best score (0.902) is obtained with SERA-DIS-NP-10 using $\mathcal{I}=30,000$ for TAC2008 while the best one (0.957) for TAC2009 is obtained with GeSERA-DIS-10 with $\mathcal{I}=10,000$.
     We thus use in Subsection~\ref{subsec:main_results} an index size $\bm{\mathcal{I}=30,000}$ and $\bm{\mathcal{I}=10,000}$ for TAC2008 and TAC2009, respectively. Surprisingly, the worst scores are obtained with $\mathcal{I}=1,778,742$, the largest and more diversified index size corresponding to all documents from our Wikipedia corpus. 
     
     
     
    
    
    \paragraph*{AQUAINT-2 Index -} 
    
    Figures~\ref{fig:pearson_pyramid_aquaint_wikipedia}-a and~\ref{fig:pearson_pyramid_aquaint_wikipedia}-c show the Pearson correlation coefficients of SERA and GeSERA with Pyramid when indexing different values of $\mathcal{I}$ from the AQUAINT-2 dataset, and when using TAC2008 and TAC2009 as query datasets. Similarly to Wikipedia, figures show that overall, the best results are obtained with small index sizes. The best score (0.928) was obtained with GeSERA-5 using $\mathcal{I}=15,000$ for TAC2008, while the best one (0.947) for TAC2009 is obtained with both SERA-DIS-NP-10 and GeSERA-DIS-5 using $\mathcal{I}=179,520$. Note that results with $\mathcal{I}=10,000$ are comparable with those obtained with $\mathcal{I}=179,520$ for TAC2009. We thus use in Subsection~\ref{subsec:main_results} an index size $\bm{\mathcal{I}=15,000}$ and $\bm{\mathcal{I}=179,520}$ for TAC2008 and TAC2009, respectively. Once again, the lowest results are obtained with the full AQUAINT-2 corpus corresponding to $\mathcal{I}=825,148$.

    \subsection{Comparison of GeSERA with the SoTA}
    \label{subsec:main_results}
    
    
    We use the best index sizes from Subsection~\ref{subsec:impact_index} to report the detailed results with the best variants of each method from Subsection~\ref{subsec:baselines}. More results are reported in the supplementary material.
    
    \subsubsection{TAC2008 query dataset} Table~\ref{tab:results_TAC2008_TAC2009}-$up$  shows correlations of the best variants of SERA, GeSERA, ROUGE, SummTriver (ST) and FRESA with two manual evaluation approaches: Pyramid and Responsiveness. Note that for SERA and GeSERA, we fix the query dataset TAC2008, while we vary the index between AQUAINT-2 and Wikipedia.
    
    \paragraph*{AQUAINT-2 Index -}
    Results show that GeSERA-5 achieves the best scores among all variants of SERA and GeSERA for both Pyramid and Responsiveness. SERA-5 is the best variant of SERA for Pyramid with Pearson and Spearman, while SERA-NP-10 provides better results for Pyramid with Kendall, and Responsiveness with all correlation measures. GeSERA-5 surpasses the two best variants of SERA by 0.015 (Pearson), 0.016 (Spearman), and 0.021 (Kendall) points for Pyramid, and by 0.02, 0.027, and 0.031 points for Responsiveness. The gains are higher for Responsiveness, and for Kendall correlations.
    
    \begin{table}[ht!]
\centering
\resizebox{0.48\textwidth}{!}
{
\begin{tabular}{|c|c|c|c||c|c|c|}
\hline
\multirow{2}{*}{\makecell{\textbf{TAC2008}}} & \multicolumn{3}{c||}{Pyramid} & \multicolumn{3}{c|}{Responsiveness} \\ \cline{2-7}
& Pearson & Spearman & Kendall & Pearson & Spearman & Kendall\\
\hline
 ST-JS-$\mathcal{T}_m$  & -0.889 & -0.827 & -0.643 & -0.820 & -0.801 & -0.608\\
\hline
 FRESA-1 & -0.487 & -0.638 & -0.537 & -0.385 & -0.498 & -0.371 \\
FRESA-4 & 0.544 & 0.257 & 0.168 & 0.596 & 0.416 & 0.296\\
\hline
 ROUGE-2-R     & \textbf{{\color{red}0.946}} & \textbf{{\color{red}0.967}} & \textbf{{\color{red}0.851}} & \textbf{\color{blue}0.894} & \textbf{\color{blue}0.918} & \textbf{\color{blue}0.755}\\
 ROUGE-3-F     & \textbf{\color{blue}0.941} & \textbf{\color{blue}0.951} & \textbf{\color{blue}0.810}  & \textbf{{\color{red}0.915}} & \textbf{{\color{red}0.924}} & \textbf{{\color{red}0.767}}\\
\hline
& \multicolumn{6}{c|}{AQUAINT-2 index ($\mathcal{I}=15,000$)} \\ \hline
SERA-5     & 0.913 & 0.908 & 0.732 & 0.845 & 0.821 & 0.624 \\
 SERA-NP-10 & 0.908 & 0.905 & 0.739 & 0.849 & 0.827 & 0.632 \\
\hline
 GeSERA-5  & \textbf{0.928} & \textbf{0.924} & \textbf{0.760} & \textbf{0.869} & \textbf{0.854} & \textbf{0.663} \\
\hline
&\multicolumn{6}{c|}{Wikipedia index ($\mathcal{I}=30,000$)} \\
\hline
SERA-NP-5   & 0.900 & 0.898 & 0.733 & 0.839 & 0.819 & 0.616 \\
SERA-DIS-NP-10 & 0.902 & 0.917 & 0.754 & 0.826 & 0.820 & 0.626 \\
\hline
GeSERA-10  & 0.883 & 0.903 & 0.727 & 0.808 & 0.805 & 0.598 \\
GeSERA-DIS-10  & 0.882 & 0.899 & 0.722 & 0.810 & 0.800 & 0.601  \\


\hline \hline
\multirow{2}{*}{\makecell{\textbf{TAC2009}}} & \multicolumn{3}{c||}{Pyramid} & \multicolumn{3}{c|}{Responsiveness} \\ \cline{2-7}
& Pearson & Spearman & Kendall & Pearson & Spearman & Kendall\\
\hline
 ST-JS-$\mathcal{T}_m$  & -0.526& -0.755 & -0.623 & -0.650 & -0.744 & -0.587 \\
\hline
FRESA-1 & -0.610 & -0.650 & -0.491 & -0.594 & -0.565 & -0.410 \\
FRESA-2 & \textbf-0.630 & 0.046 & -0.026 & -0.385 & -0.074 & -0.063 \\
\hline

 ROUGE-1-F     & \textbf{0.951} & {\textbf{\color{blue}0.915}} & \textbf{0.788}  & \textbf{0.835} & 0.793 & 0.622\\
 ROUGE-3-F     & 0.842 & \textbf{{\color{red}0.964}} & \textbf{{\color{blue}0.841}}  & 0.622 & \textbf{{\color{red}0.852}} & \textbf{{\color{red}0.675}} \\
 ROUGE-3-R     & 0.848 & \textbf{{\color{red}0.964}} & \textbf{{\color{red}0.845}} & 0.627 & \textbf{{\color{blue}0.845}} & {\textbf{\color{blue}0.673}} \\
\hline
& \multicolumn{6}{c|}{AQUAINT-2 index ($\mathcal{I}=179,520$)} \\ \hline
SERA-NP-5     & 0.916 & 0.831 & 0.670 & 0.816 & 0.692 & 0.530 \\
 SERA-NP-10    & 0.885 & 0.828 & 0.662 & 0.806 & 0.702 & 0.529 \\
 SERA-DIS-10   & 0.941 & 0.836 & 0.671 &  0.806 & 0.673 & 0.521 \\
 SERA-DIS-NP-10 & 0.947 & 0.825 & 0.665 &  0.806 & 0.683 & 0.518 \\
\hline
GeSERA-5   & 0.908 & 0.835 & 0.678 & 0.834 & 0.697 & 0.530  \\
GeSERA-DIS-5 & 0.947 & 0.836 & 0.688 & 0.831 & 0.684 & 0.525 \\
\hline
&\multicolumn{6}{c|}{Wikipedia index ($\mathcal{I}=10,000$)} \\ 
\hline
SERA-10           & 0.944     & 0.892     & 0.741     & \textbf{\color{red}0.845} & \textbf{0.784}   & \textbf{0.607}  \\
SERA-KW-10        & 0.944     & 0.894 & 0.738  & 0.839  & 0.771 & 0.588   \\
SERA-DIS-10       & \textbf{{\color{blue} 0.955}} & 0.896 & 0.751   & 0.791  & 0.781            & 0.603 \\
 SERA-DIS-KW-10    & 0.952     & \textbf{0.899} & 0.753 & 0.785  & 0.782 & \textbf{0.607}  \\
\hline
GeSERA-10       & 0.935  & 0.870  & 0.710 & \textbf{{\color{blue}0.839}} & 0.737    & 0.571 \\
 GeSERA-DIS-5    & 0.952  & 0.867  & 0.717  & 0.819   & 0.768   & 0.592 \\
 GeSERA-DIS-10   & \textbf{{\color{red}0.957}} & 0.882 & 0.710   & 0.800   & 0.748   & 0.577 \\
\hline 
\end{tabular}
}
\caption{Correlations on TAC2008 and TAC2009 datasets, in terms of Pearson, Spearman and Kendall, of automatic evaluation methods with Pyramid and Responsiveness. The best first (red), second (blue) and third (black) scores of each column are in bold}
\label{tab:results_TAC2008_TAC2009}
\end{table}

    \vspace{-0.5em}
    
    \paragraph*{Wikipedia Index -}
    Results show that GeSERA-10 is the best variant of GeSERA for Pyramid with all correlation metrics used. 
    GeSERA-DIS-10 gets the best scores for Responsiveness with Pearson and Kendall. Alternatively, the best SERA variant is SERA-DIS-NP-10 for Pyramid with all correlation measures, and with Responsiveness for Spearman and Kendall measures. Interestingly, when using queries from TAC2008 with Wikipedia, GeSERA does not surpass SERA neither for Pyramid, not for Responsiveness.


    
    While ROUGE-2-R and ROUGE-3-F provide the best results for all correlation measures on TAC2008, GeSERA and SERA largely surpass the scores of SummTriver and FRESA with both Pyramid and Responsiveness. In the case of GeSERA-5, it achieves higher correlations than ST-JS-$\mathcal{T}_m$, the best variant of SummTriver, by 0.039, 0.097, and 0.117 for Pyramid, and 0.049, 0.053, and 0.055 for Responsiveness. Finally, FRESA baseline achieves the lowest correlation scores in all configurations. The performance of SummTriver and FRESA is not surprising insofar as they do not rely on any human reference.

    \subsubsection{TAC2009 query dataset} Table~\ref{tab:results_TAC2008_TAC2009}-$bottom$ shows correlation coefficients of SERA, GeSERA, ROUGE, SummTriver and FRESA with two manual evaluation approaches: Pyramid and Responsiveness. Once again, we fix the query dataset TAC2009, while we vary the index between AQUAINT-2 and Wikipedia.
    
     \paragraph*{AQUAINT-2 Index -}
        ROUGE provides the highest scores against SERA and GeSERA when we index documents from AQUAINT-2. Importantly, GeSERA-DIS-5 and GeSERA-5 achieve higher correlations than SERA with Pyramid and Responsiveness, respectively. 
        Note that the scores of SERA vary more between its variants, while results of GeSERA are more stable and the best ones are obtained with only two of its variants. This finding highlights the robustness of our approach against variations of configurations.
        
    
         \paragraph*{Wikipedia Index -}
    
        SERA and GeSERA surpass ROUGE against the Pyramid manual method in terms of Pearson correlation when indexing documents from Wikipedia.
        The best correlations are obtained by GeSERA-DIS-10 and SERA-10 against Pyramid and Responsiveness, respectively. 
    
        Interestingly, for TAC2009, GeSERA-DIS-10 achieves better Pearson correlation than ROUGE with Pyramid, and GeSERA-10 with Responsiveness. This finding proves the effectiveness of GeSERA to evaluate summaries from the general domain. Equally, GeSERA reduces the gap between SERA and ROUGE in most of other cases.

    SummTriver achieves reasonably good results in Table~\ref{tab:results_TAC2008_TAC2009} even without the use of any human reference. This baseline is useful when human summaries are costly or hard to find. However, when such references are available, SummTriver does not take advantage of them, leading its correlation to be low compared to human-based evaluation approaches such as ROUGE and SERA.

    FRESA shows the lowest scores among evaluation approaches tested here. It drops approximately from 0.1 to 0.3 point compared to the lowest results obtained by SERA. This is mainly because FRESA is based only on the divergence between the evaluated summary and its source documents, without including any comparison with summaries generated by other participants, as Summtriver does. Thus, FRESA is barely correlated with manual evaluation in many cases where the correlation gets so close to zero (for instance, FRESA-2 with TAC2009 using Kendall correlation). Note that SummTriver and FRESA have mostly negative correlations because they are based on a divergence measure which increases when the summary's quality is low and decreases when its quality is high.

    


    \subsubsection{CNNDM query dataset}
    \label{subsubsec:cnn_db}
    
    Based on results obtained in Subsection~\ref{subsec:impact_index} regarding the effectiveness of SERA and GeSERA with small index sizes, we decided to index $\mathcal{I}=10,000$ documents from Wikipedia to run the two methods on CNNDM. Results are in Table \ref{tab:results_CNNDM}.
    
      \begin{table}[hb!]
      \centering
        \resizebox{0.48\textwidth}{!}
        {
        \begin{tabular}{|c|c|c|c|}
        \hline
         \textbf{CNNDM} & Pearson & Spearman & Kendall\\
        \hline
         ROUGE-1-R  & {\textbf{\color{blue}0.914}} & {\textbf{\color{blue}0.922}} & {\textbf{\color{blue}0.773}}  \\
         ROUGE-2-R  & \textbf{\color{red}0.962} & \textbf{\color{red}0.958} & \textbf{\color{red}0.860}\\
         ROUGE-L-F  & 0.526 & 0.368 & 0.278 \\
        \hline
         BERTScore-1-P  & -0.021 & 0.093 & 0.064  \\
         BERTScore-1-R  & 0.768 & 0.738 & 0.552  \\
        \hline
         MoverScore  & 0.443 & 0.367 & 0.284 \\
        \hline
         JS-2  & 0.780 & 0.665 & 0.512 \\
        \hline
         SERA-10    & 0.858 & 0.789 & 0.616\\
         SERA-KW-10 & 0.864 & 0.782 & 0.621 \\
         SERA-DIS-10 & 0.827 & 0.781 & 0.605 \\
         SERA-DIS-NP-5  & 0.599 & 0.554 & 0.391 \\
        \hline
         GeSERA-5 & 0.623 & 0.527 & 0.387  \\
         GeSERA-10 & \textbf{0.880} & \textbf{0.872} & \textbf{0.719} \\
         GeSERA-DIS-10 & 0.817 & 0.788 & 0.605 \\
        \hline
        \end{tabular}
        }
    \caption{Correlation coefficients on CNNDM, in terms of Pearson, Spearman and Kendall, of multiple automatic evaluation methods with LitePyramid.}
    \label{tab:results_CNNDM}
    \end{table}

    Table~\ref{tab:results_CNNDM} shows that the highest correlations of ROUGE are obtained with ROUGE-2-R, followed by ROUGE-1-R. Globally, the highest correlations in ROUGE are obtained with the recall metric (ROUGE-R), followed by ROUGE-F, and finally by ROUGE-P. The following highest correlations are obtained with GeSERA-10. Once again, GeSERA surpasses all the SERA variants, and the state-of-the-art methods presented in this table. 
    
    Although SERA-KW-10 has the best score in terms of Pearson and Kendall, all SERA variants present very similar scores. Behind the SERA method, BERTScore and JS-2 measures present very similar scores. Meanwhile, MoverScore shows the lowest correlations. Results show the effectiveness of GeSERA to evaluate extractive and abstractive summaries since CNNDM contains both approaches.

\subsection{Impact of human annotators on the performance of SERA and GeSERA}
    \label{subsec:impact_human_annotators}
    
    \begin{figure*}[ht!] 
   \begin{minipage}{0.5\linewidth}
    \centering
    \includegraphics[width=\linewidth]{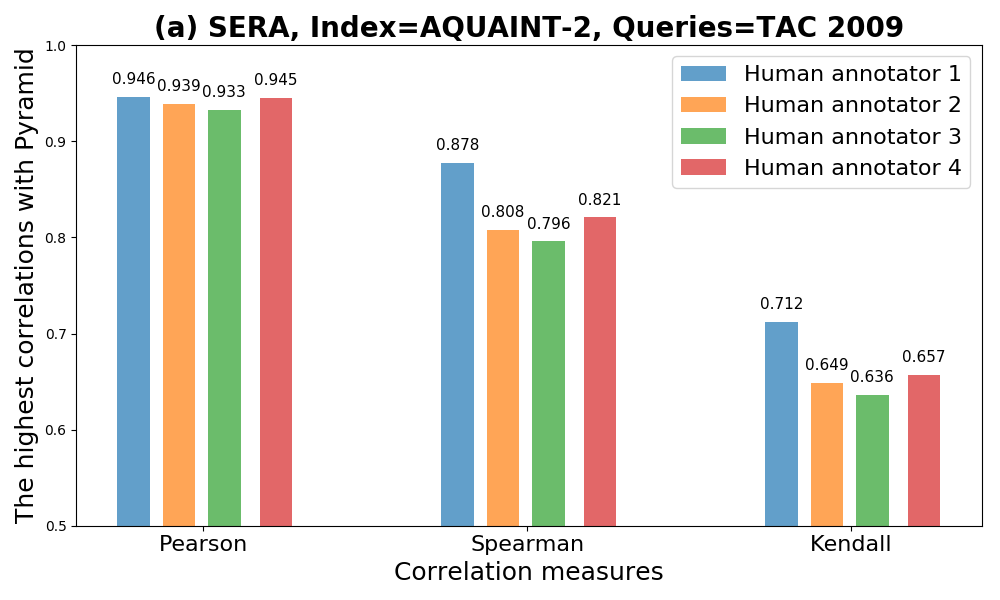} 
  \end{minipage}\hfill
    \begin{minipage}{0.5\linewidth}
    \centering
    \includegraphics[width=\linewidth]{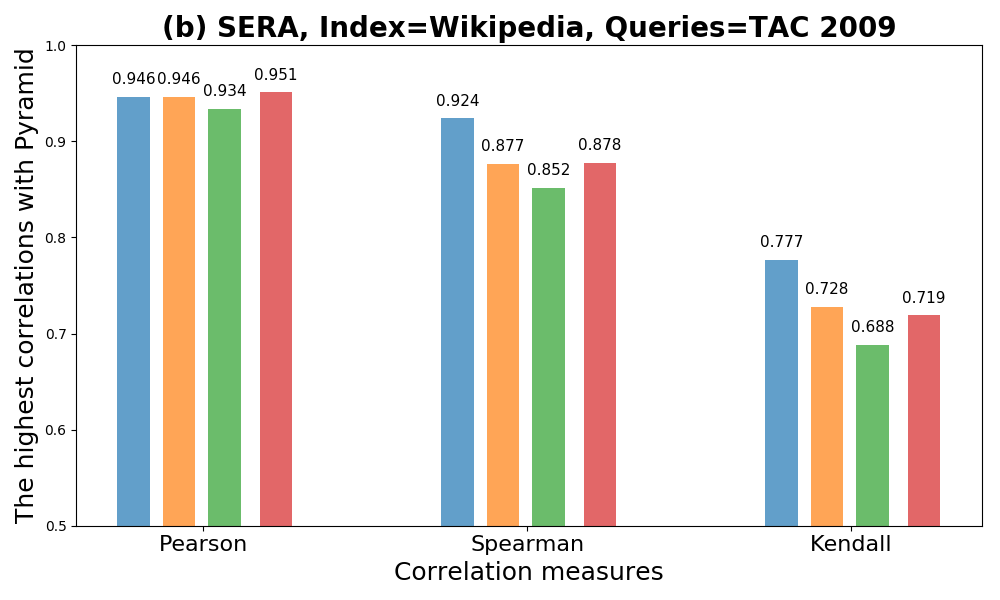} 
  \end{minipage}\hfill
    \begin{minipage}{0.5\linewidth}
    \centering
    \includegraphics[width=\linewidth]{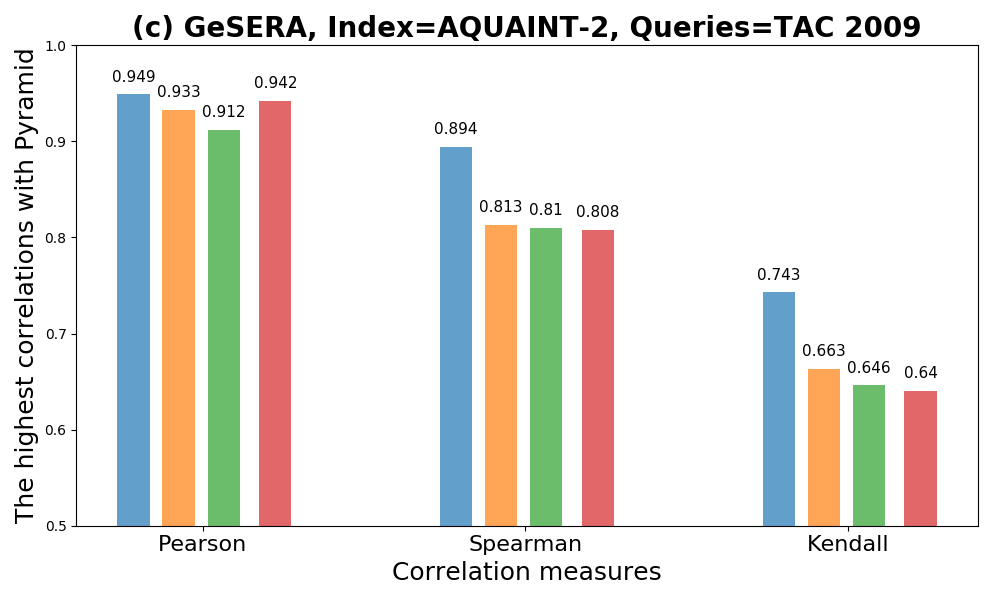} 
  \end{minipage}\hfill
    \begin{minipage}{0.5\linewidth}
    \centering
    \includegraphics[width=\linewidth]{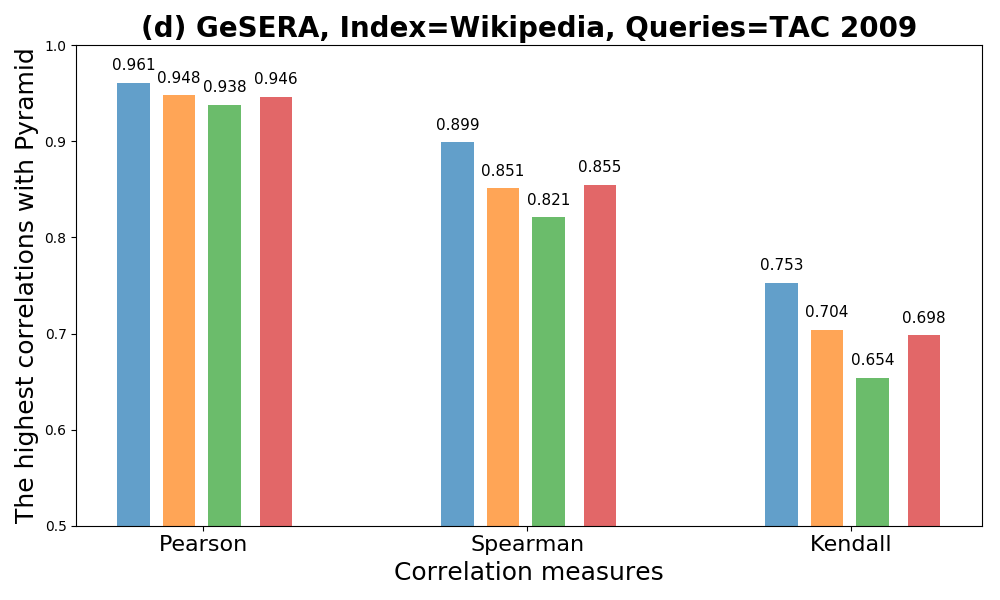} 
  \end{minipage}\hfill
  \caption{Correlation coefficients in terms of Pearson, Spearman and Kendall obtained by each annotator $\mathcal{A}_i$ using TAC2009 for queries, and AQUAINT-2 and Wikipedia as indexes. \textit{Best viewed in color.}}
  \label{fig:annotators} 
\end{figure*}

    For the sake of comparability with state-of-the-art approaches, we presented in the previous section correlations computed with the four manual annotators. According to~\citet{lin-hovy-2002-manual}, human evaluation is subjective. We confirm experimentally this finding and highlight that human annotators affect the performance of automatic evaluation approaches. To know how much each annotator can affect the correlation against human evaluations, and which annotator gets the lowest and highest correlations, we compute scores using different combination of human annotators. We compute the correlation of each human annotator individually, using three human annotators: ($\mathcal{A}_1$, $\mathcal{A}_2$, $\mathcal{A}_3$), ($\mathcal{A}_1$, $\mathcal{A}_2$, $\mathcal{A}_4$), ($\mathcal{A}_2$, $\mathcal{A}_3$, $\mathcal{A}_4$), and finally using the four human annotators ($\mathcal{A}_1$, $\mathcal{A}_2$, $\mathcal{A}_3$, $\mathcal{A}_4$). 
    
    Note that we only report here the results on TAC2009 dataset insofar as the results on TAC2008 are not conclusive, where the best score vary considerably from one annotator to another depending on the configuration. We present results on TAC2008 in the supplementary material.
    


    Figure~\ref{fig:annotators} provides SERA and GeSERA correlations with Pyramid using TAC2009 as a query dataset and AQUAINT-2 and Wikipedia as indexes. Results show that the best human annotator is always $\mathcal{A}_1$ as he provides summaries with the best correlations of SERA and GeSERA in terms of Pearson, Spearman, and Kendall. 
    Inversely, the worst human annotator is always $\mathcal{A}_3$ for SERA as he achieves the worst scores in terms of all correlation metrics used here. For GeSERA, the worst human annotator is $\mathcal{A}_3$ for Wikipedia in terms of all correlation metrics, while it is $\mathcal{A}_4$ for AQUAINT-2 in terms of Spearman and Kendall and $\mathcal{A}_3$ in terms of Pearson.  
    
    \begin{table}[htb!]
      \centering
       \resizebox{0.48\textwidth}{!}
    {
        \begin{tabular}{|c||c|c||c|c||c|c||c|c|}
        \cline{2-9}
        \multicolumn{1}{c|}{} & \multicolumn{4}{c||}{AQUAINT-2} & \multicolumn{4}{c|}{Wikipedia} \\ \cline{2-9} 
        \multicolumn{1}{c|}{} & \multicolumn{2}{c||}{\small SERA-DIS-NP-10}  & \multicolumn{2}{c||}{\small GeSERA-DIS-5}  &\multicolumn{2}{c||}{\small SERA-DIS-10}   &\multicolumn{2}{c|}{\small GeSERA-DIS-10}  \\ \hline
        Annotators  & 4 & 3 & 4 & 3 & 4 & 3 & 4 & 3\\ \hline
        Pearson  & 0.947 & \textbf{0.949}  & 0.947 & \textbf{0.951} & 0.955 & \textbf{0.959}  & 0.957 & \textbf{0.959}   \\ \hline
        Spearman & 0.825 & \textbf{0.835}  & 0.836 & \textbf{0.841} & 0.896 & \textbf{0.909}  & \textbf{0.882} & \textbf{0.882}   \\ \hline
        Kendall  & 0.665 & \textbf{0.681} & 0.668 & \textbf{0.691} & 0.751 & \textbf{0.776}  & 0.710 & \textbf{0.743}   \\ \hline
        \end{tabular}
        }
      \caption{Impact of human annotators on the evaluation with SERA and GeSERA using TAC2009.}
      \label{tab:annotators}
    \end{table}
    
    In Table~\ref{tab:annotators}, we compare results obtained with the four manual annotators versus those obtained with the best three annotators ($\mathcal{A}_1$, $\mathcal{A}_2$, $\mathcal{A}_4$) for TAC2009. Results show that there is a clear gain when discarding the most unreliable annotator. We conclude that human annotators partially participate in the quality of automatic summary evaluation. This bias is caused by the quality of their manually written summaries.

    \section{Related work}
    \label{sec:related_work}
    
    Evaluation methods are fundamental techniques to assess if summaries generated by an automatic system capture the original document’s idea. Different evaluation methods have been developed in the last decade for the evaluation of automatically-generated summaries. There exist two types of evaluation methods: (1) manual evaluation methods, and (2) automatic evaluation methods. The first group of methods requires human intervention as ground-truth references. Pyramid~\citep{NenkovaP04} and Responsiveness are the most popular such methods. The second group of methods is divided itself into two subsets: (1) methods that need human intervention like ROUGE~\cite{Lin04rougea} and SERA~\cite{Cohan2016}, and (2) methods that do not need any human reference like SummTriver~\cite{CabreraDiego2018_SummTriver} and FRESA~\cite{Torres2010_FRESA}.
    
    
    The most popular automatic metric used by the community is ROUGE~\citep{Lin04rougea}. It needs  reference summaries, and is based on their lexical overlaps with candidate summaries. That is why it is more useful to evaluate extractive summaries where chunks of the text are copied and pasted to form the summary. However, in the case of abstractive summaries where the ATS paraphrases the text with possibly new vocabulary, the ROUGE metric becomes unfair. To overcome this issue, researchers have been proposing in the last few years other automatic metrics to fairly evaluate both extractive and abstractive summaries. 
    
    The first type of automatic evaluation methods relies partially on human judgment as ROUGE does. The simplest method is based on a  \textit{Jensen-Shannon} (JS-2)~\citep{lin2006_information} divergence between bi-gram's distribution of candidate and reference summaries. More sophisticated systems include MoverScore~\cite{zhao2019_moverScore} that is based on fine-tuning the BERT model and combining contextualized representations with Earth Mover Distance (EMD) from~\citet{rubner2000_earthMoverDistance}. BERTScore~\cite{Zhang2020_BERTScore} also is based on BERT model. Unlike ROUGE~\citep{Lin2004_rougea}, BERTScore makes use of contextual embeddings that are effective for paraphrase detection. Similarly to BERTScore, \textit{Semantic Similarity for Abstractive Summarization} (SSAS)~\cite{vadapalli2017_ssas} is based on semantic matching between candidate and reference summaries.  The second type of automatic evaluation methods does not need any human intervention.  For instance,   \textit{FRamework for Evaluating Summaries Automatically} (FRESA)~\cite{Torres2010_FRESA} is based on divergences among probability distributions between the summary to evaluate and its source document. Another well-known metric is SummTriver~\cite{CabreraDiego2018_SummTriver}. It is based on Trivergences between the summary to evaluate, its source document(s), and a set of summaries related to the same source document(s) but generated with other ATS systems.
    
   \vspace{-0.75em}
    
    \section{Conclusion and perspectives}
    \label{sec:conclusion}

    We introduced GeSERA, an open-source system for general-domain summary evaluation. We redefine query reformulation of SERA based on POS Tags analysis of datasets from different domains, and replace the biomedical index with documents from AQUAINT-2 and Wikipedia. GeSERA achieves competitive results compared to state-of-the-art approaches. Overall, GeSERA surpasses SERA and reduces its gap with ROUGE, and in two cases, it even surpasses ROUGE, the lexical-based method. Unsurprisingly, the comparison with evaluation methods that do not rely on human references reveals a large gap in favor of GeSERA since it relies on human references while the others do not. Extensive experiments show that the index size has a considerable effect on the performance of SERA and GeSERA that tend to perform better with small-size indexes. Finally, the study of human annotators shows their impact on the performance of automatic evaluation methods that rely on human intervention. Our code is publicly available to facilitate reproducibility. We will:
(1) explore other variants of the search engine to know its impact on GeSERA,
(2) propose a new version of GeSERA that does not rely on human intervention by exploiting information from the source text,
(3) apply prepossessing on the index and search for other solutions to improve query reformulation, and (4) explore larger query datasets such as Multi-News \citep{fabbri2019_multinews}.


\clearpage
\newpage



\bibliographystyle{acl_natbib}
\bibliography{anthology,ranlp2021}

\clearpage \newpage

\appendix

\section*{Supplementary material}
In this supplementary material, we provide: (1) more details about evaluation datasets (Section \ref{sec:datasets_supp}), (2) implementation details (Section \ref{sec:implementation_supp}), (3) detailed results of all tested approaches (Section \ref{sec:results_supp}), and (4) the impact of human annotators on TAC2008 dataset (Section \ref{subsec:impact_human_annotators_supp})
    

\section{Datasets details}
\label{sec:datasets_supp}

\begin{itemize}[leftmargin=*]
\setlength\itemsep{-0.3em}

\item \textbf{AQUAINT-2} is a news corpus built from New York Times, Associated Press, and Xinhua News Agency. Indexes built from this corpus are balanced, except for the largest one ($\mathcal{I}=825,148$), that contains all documents.  Note that AQUAINT-2 is not open-source, and we cannot distribute it. However, obtained results can be helpful in academic research.

\item \textbf{Wikipedia} is a free online encyclopedia from the general domain. The largest index ($\mathcal{I}=1,778,742$) contains all available documents. 
    
\item \textbf{TAC2008 (/TAC2009)} contains two sets. Each set contains 48 (/44) topics. Each topic includes 10 documents
and 4 reference summaries. Candidate summaries are proposed by 58 (/55) participants, where each one provides a candidate summary per topic. In total, there are 960 (/880) documents, 5568 (/4840) candidate summaries, and 384 (/352) reference summaries.
\end{itemize}

    

\section{Implementation details}
\label{sec:implementation_supp}

    SERA and GeSERA were implemented in Python. For information retrieval, we used the Okapi BM25F ranking function from Whoosh, a flexible and pure python search engine framework. 

    We used the authors' public implementations to run ROUGE, SummTriver, and FRESA. 
    The latter was basically designed for mono-document evaluation. Thus, we concatenated all the articles of the same topic to be able to run it on TAC.
    
    
    To compute the correlations with LitePyramid of ROUGE, BERTScore, MoverScore, and JS-2 on the CNN Daily Mail dataset, we used the scores provided by~{\color{blue}Bhandari et al., 2020} in their GitHub repository\footnote{ \url{https://github.com/neulab/REALSumm/}}. Based on experiments on the TAC datasets in Subsection~{\color{blue}4.3}, we use an index size of $\mathcal{I}=10,000$ in SERA and GeSERA.

    For the sake of comparability, scores are averaged for each participant before computing the correlations with manual methods.
    

\section{More results}
\label{sec:results_supp}
Table~\ref{tab:results_CNNDM_supp} and Table~\ref{tab:results_TAC_supp} provide more results on CNNDM, TAC2008 and TAC2009 datasets. 
Both tables present variants of the evaluation metrics that we did not report in the main paper.




    \begin{table}[htb!]
      \centering
        \resizebox{0.48\textwidth}{!}
        {
        \begin{tabular}{|c|c|c|c|}
        \hline
         & Pearson & Spearman & Kendall\\
        \hline
         ROUGE-1-F  & 0.600 & 0.468 & 0.358 \\
        ROUGE-1-P  & -0.175 & -0.212 & -0.117  \\
        ROUGE-2-F  & 0.648 & 0.452 & 0.311  \\
        ROUGE-2-P  & 0.099 & 0.050 & 0.023  \\
        ROUGE-L-P  & -0.045 & -0.148 & -0.070 \\
        ROUGE-L-R  & 0.871 & -0.914 & 0.759 \\
        \hline
         BERTScore-1-F  & 0.385 & 0.374 & 0.258 \\
        \hline
        MoverScore  & 0.443 & 0.367 & 0.284 \\
        \hline
        JS-2  & 0.780 & 0.665 & 0.512\\
        \hline
        SERA-5     & 0.773 & 0.710 & 0.508 \\
         SERA-NP-5  & 0.690 & 0.639 & 0.452  \\
         SERA-NP-10 & 0.743 & 0.705 & 0.502  \\
         SERA-KW-5  & 0.784 & 0.711 & 0.508  \\
         SERA-DIS-5 & 0.748 & 0.668 & 0.472 \\
         SERA-DIS-NP-10 & 0.671 & 0.568 & 0.393  \\
         SERA-DIS-KW-5  & 0.758 & 0.657 & 0.465  \\
         SERA-DIS-KW-10 & 0.828 & 0.752 & 0.565  \\
        \hline
         GeSERA-DIS-5 & 0.566 & 0.469 & 0.315  \\
        \hline
        \end{tabular}
        }
    \caption{Correlations of CNNDM dataset, in terms of Pearson, Spearman and Kendall, of multiple automatic evaluation methods with LitePyramid. 
    }
    \label{tab:results_CNNDM_supp}
    \end{table}



\section{Impact of human annotators}
\label{subsec:impact_human_annotators_supp}

\begin{figure}[hb!] 
 
 \begin{subfigure}[b]{0.49\textwidth}
   \includegraphics[width=1\linewidth]{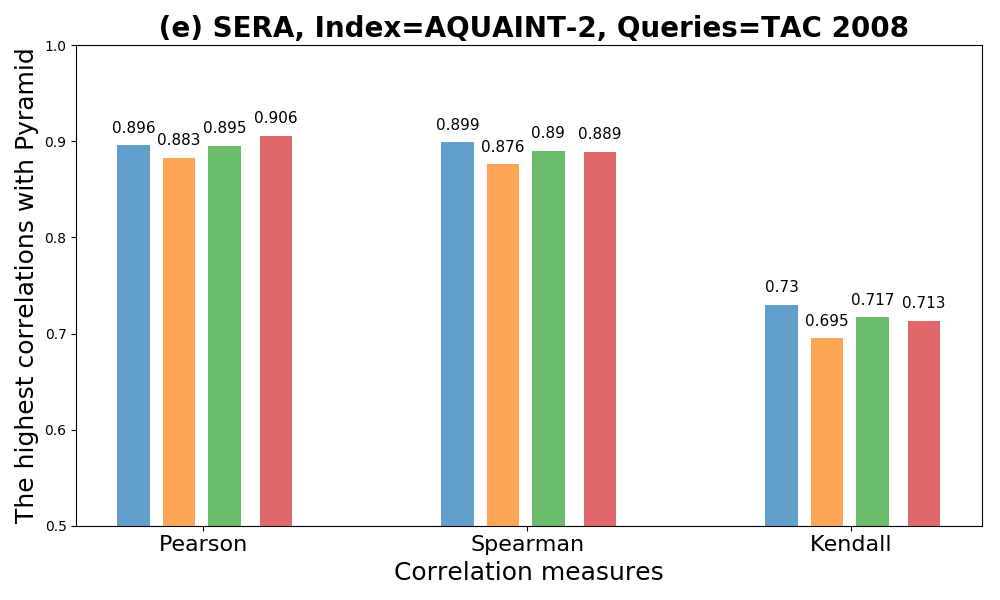}
\end{subfigure}

 \begin{subfigure}[b]{0.49\textwidth}
   \includegraphics[width=1\linewidth]{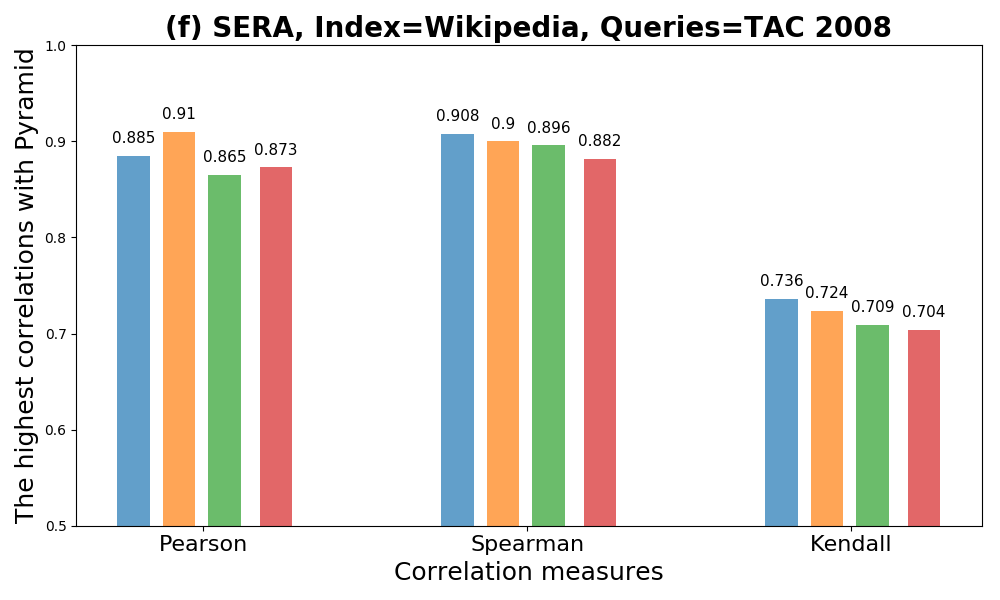}
\end{subfigure}

 \begin{subfigure}[b]{0.49\textwidth}
   \includegraphics[width=1\linewidth]{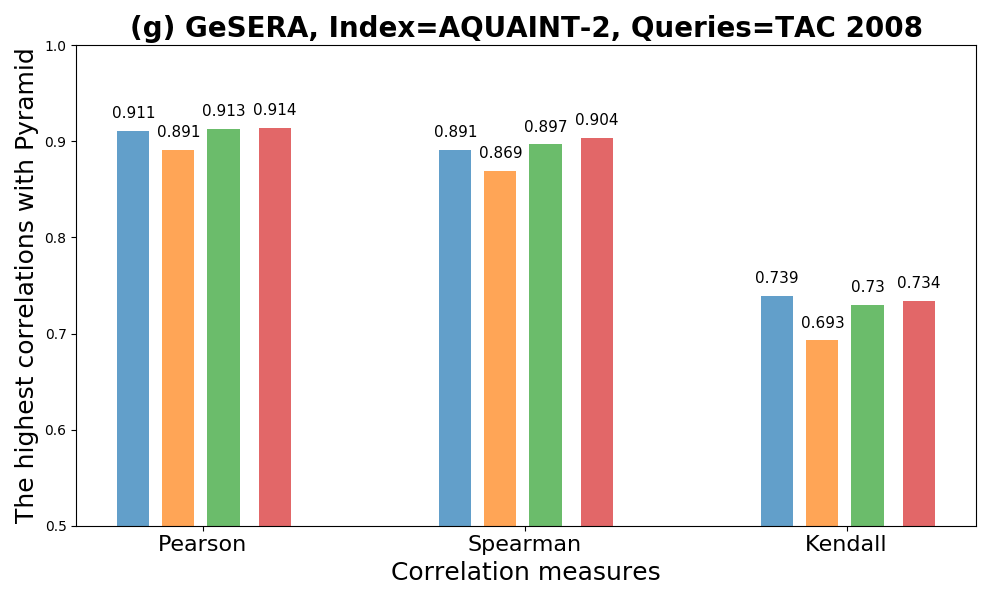}
\end{subfigure}

 \begin{subfigure}[b]{0.49\textwidth}
   \includegraphics[width=1\linewidth]{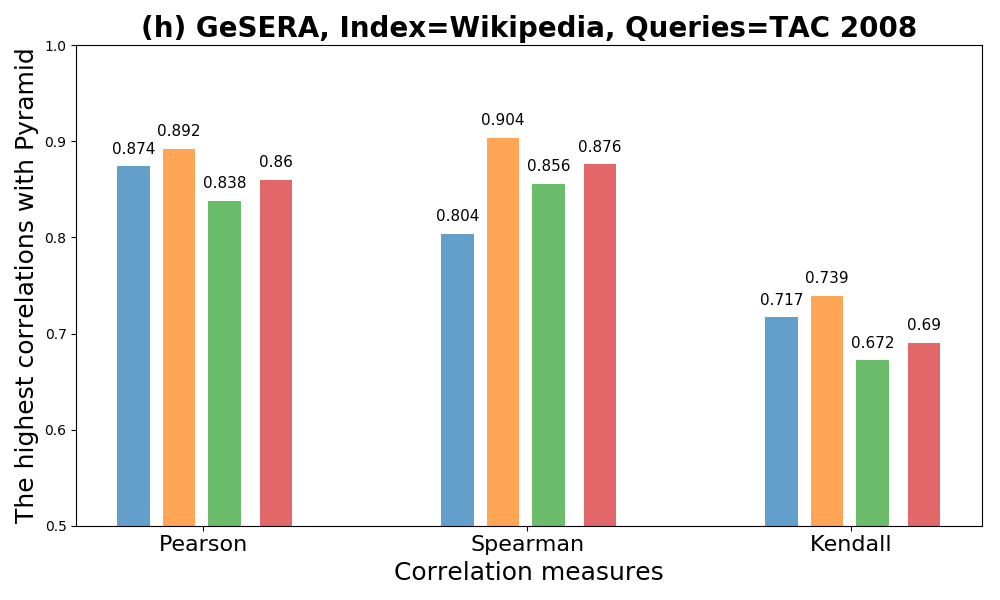}
\end{subfigure}
  \caption{Correlations coefficients obtained by each annotator $\mathcal{A}_i$ using TAC2008 dataset for queries, and AQUAINT-2 and Wikipedia as indexes.}
  \label{fig:annotators_supp} 
\end{figure}

 Figure~\ref{fig:annotators_supp} provides SERA and GeSERA correlations with Pyramid using TAC2008 as a query dataset and AQUAINT-2 and Wikipedia as indexes.
Contrarily to TAC2009, it is hard to define for TAC2008 the impact of human annotators on the evaluation with SERA and GeSERA, as the best scores change from one case to another. For instance, $\mathcal{A}_1$ is the best human annotator for SERA with both AQUAINT-2 and Wikipedia corpora in terms of Spearman and Kendall. However, in terms of Pearson correlation, the best annotator is $\mathcal{A}_4$ for AQUAINT-2 and $\mathcal{A}_2$ for Wikipedia. Alternatively, the best annotator for GeSERA is always $\mathcal{A}_2$ for Wikipedia while the same annotator provides the worst results with AQUAINT-

\begin{table}[ht!]
\centering
\resizebox{0.48\textwidth}{!}
{
\begin{tabular}{|c|c|c|c||c|c|c|}
\hline
\multirow{3}{*}{} & \multicolumn{6}{c||}{TAC2008}  \\ \cline{2-7}
\multirow{3}{*}{} & \multicolumn{3}{c||}{Pyramid} & \multicolumn{3}{c||}{Responsiveness} \\ \cline{2-7}
& Pearson & Spearman & Kendall & Pearson & Spearman & Kendall\\
\hline
ST-sJS-$\mathcal{T}_m$ & -0.885 & -0.822 & -0.637 & -0.822 & -0.797 & -0.605 \\
 ST-KL-$\mathcal{T}_m$ & -0.694 & -0.700 & -0.510 & -0.706 & -0.695 & -0.504 \\
 ST-JS-$\mathcal{T}_c$ & -0.858 & -0.805 & -0.613 & -0.771 & -0.777 & -0.578 \\
 ST-sJS-$\mathcal{T}_c$ & -0.857 & -0.805 & -0.612 & -0.771 & -0.777 & -0.577 \\
 ST-KL-$\mathcal{T}_c$ & -0.216 & -0.168 & -0.123 & 0.025 & 0.134 & 0.091  \\
\hline
FRESA-2 & 0.474 & -0.062 & -0.064 & 0.523 & 0.076 & 0.034 \\
FRESA-3 & 0.539 & 0.241 & 0.162 & 0.593 & 0.362 & 0.250 \\
\hline
 ROUGE-1-F     & 0.908 & 0.941 & 0.787  & 0.853 & 0.883 & 0.702 \\
 ROUGE-1-P     & 0.730 & 0.841 & 0.643  & 0.698 & 0.803 & 0.626 \\
 ROUGE-1-R     & 0.911 & 0.935 & 0.774  & 0.851 & 0.858 & 0.665 \\
 ROUGE-2-F     & 0.940 & 0.965 & 0.843 & 0.892 & 0.915 & 0.746 \\
 ROUGE-2-P     & 0.911 & 0.942 & 0.788 & 0.873 & 0.901 & 0.730\\
 ROUGE-3-P     & 0.926 & 0.934 & 0.783 & 0.909 & 0.918 & 0.766 \\
 ROUGE-3-R     & 0.945 & 0.951 & 0.811  & 0.914 & 0.922 & 0.763\\
 ROUGE-L-F     & 0.878 & 0.925 & 0.756  & 0.823 & 0.868 & 0.689\\
 ROUGE-L-P     & 0.711 & 0.823 & 0.632  & 0.679 & 0.794 & 0.611\\
 ROUGE-L-R     & 0.882 & 0.927 & 0.762  & 0.823 & 0.856 & 0.661\\
 ROUGE-W-1.2-F & 0.901 & 0.940 & 0.782  & 0.848 & 0.878 & 0.701\\
 ROUGE-W-1.2-P & 0.712 & 0.822 & 0.631 & 0.688 & 0.794 & 0.620 \\
 ROUGE-W-1.2-R & 0.897 & 0.940 & 0.785 & 0.841 & 0.871 & 0.684 \\
 ROUGE-SU4-F   & 0.917 & 0.949 & 0.805 & 0.870 & 0.904 & 0.728 \\
 ROUGE-SU4-P   & 0.839 & 0.910 & 0.728 & 0.805 & 0.869 & 0.689 \\
 ROUGE-SU4-R   & 0.927 & 0.950 & 0.800 & 0.874 & 0.908 & 0.736 \\
\hline
& \multicolumn{6}{c|}{AQUAINT-2 index ($\mathcal{I}=15,000$)} \\ \hline
 SERA-10    & 0.887 & 0.871 & 0.693 & 0.817 & 0.766 & 0.572 \\
 SERA-NP-5  & 0.866 & 0.863 & 0.681 & 0.792 & 0.770 & 0.569 \\
 SERA-KW-5  & 0.909 & 0.901 & 0.721 & 0.841 & 0.809 & 0.611 \\
 SERA-KW-10 & 0.890 & 0.880 & 0.705 & 0.823 & 0.779 & 0.579 \\
 SERA-DIS-5 & 0.905 & 0.885 & 0.713 & 0.840 & 0.800  & 0.593 \\
 SERA-DIS-10 & 0.900 & 0.888 & 0.711 & 0.829 & 0.797 & 0.592  \\
 SERA-DIS-NP-5  & 0.875 & 0.864 & 0.679 & 0.805 & 0.774 & 0.573  \\
 SERA-DIS-NP-10 & 0.907 & 0.905 & 0.735 & 0.843 & 0.819 & 0.616  \\
 SERA-DIS-KW-5  & 0.903 & 0.885 & 0.712 & 0.837 & 0.801 & 0.597  \\
 SERA-DIS-KW-10 & 0.902 & 0.888 & 0.709 & 0.832 & 0.804 & 0.601  \\
\hline
 GeSERA-10 & 0.902 & 0.890 & 0.708 & 0.843 & 0.800 & 0.604  \\
 GeSERA-DIS-5 & 0.924 & 0.910 & 0.746 & 0.861 & 0.836 & 0.641 \\
 GeSERA-DIS-10  & 0.918 & 0.896 & 0.724 & 0.852 & 0.806 & 0.610  \\ 
\hline
& \multicolumn{6}{c|}{Wikipedia index ($\mathcal{I}=30,000$)} \\ \hline
 SERA-5      & 0.831 & 0.839 & 0.673 & 0.763 & 0.751 & 0.560   \\
 SERA-10     & 0.884 & 0.900 & 0.724 & 0.812 & 0.798 & 0.594  \\
 SERA-NP-10  & 0.890 & 0.912 & 0.738 & 0.806 & 0.812 & 0.618  \\
 SERA-KW-5   & 0.837 & 0.838 & 0.667 & 0.767 & 0.749 & 0.552  \\
 SERA-KW-10  & 0.885 & 0.906 & 0.727 & 0.812 & 0.806 & 0.603   \\
 SERA-DIS-5  & 0.833 & 0.825 & 0.655 & 0.781 & 0.757 & 0.568 \\
 SERA-DIS-10 & 0.877 & 0.887 & 0.707 & 0.815 & 0.790 & 0.588 \\
 SERA-DIS-NP-5  & 0.894 & 0.884 & 0.718 & 0.838 &  0.809 & 0.604 \\
 SERA-DIS-KW-5  & 0.838 & 0.837 & 0.667 & 0.783 & 0.761 & 0.567  \\
 SERA-DIS-KW-10 & 0.881 & 0.894 & 0.719 & 0.817 & 0.797 & 0.598\\
\hline
 GeSERA-5   & 0.873 & 0.865 & 0.698 & 0.803 & 0.774 & 0.581  \\
 GeSERA-DIS-5  & 0.870 & 0.865 & 0.701 & 0.802 & 0.773 & 0.580 \\
\hline
\hline
\multirow{3}{*}{} & \multicolumn{6}{c||}{TAC2009}  \\ \cline{2-7}
\hline 
ST-sJS-$\mathcal{T}_m$ & -0.511 & -0.751 & -0.620 & -0.636 & -0.739 & -0.585 \\
ST-KL-$\mathcal{T}_m$ & -0.371 & -0.681 & -0.558 & -0.518 & -0.683 & -0.550 \\
ST-JS-$\mathcal{T}_c$ & -0.477 & -0.718 & -0.582 & -0.619 & -0.710 & -0.563 \\
ST-sJS-$\mathcal{T}_c$ & -0.475 & -0.717 & -0.581 & -0.618 & -0.709 & -0.562 \\
ST-KL-$\mathcal{T}_c$ &-0.138 & -0.062 & -0.040 & -0.014 & -0.007 & -0.005 \\
\hline 
FRESA-3 & -0.556 & 0.055 & 0.056  & -0.298 & 0.180 & -0.147\\
FRESA-4 &  -0.516 & 0.189 & 0.142  & -0.217 & 0.363 & -0.278 \\ \hline

ROUGE-1-P     & 0.923 & 0.845 & 0.678 & 0.791 & 0.789 & 0.630\\
ROUGE-1-R     & 0.926 & 0.892 & 0.748 & 0.814 & 0.764 & 0.591\\
ROUGE-2-F     & 0.930 & 0.955 & 0.839 & 0.740 & 0.831 & 0.664\\
ROUGE-2-P     & 0.906 & 0.937 & 0.796  & 0.716 & 0.829 & 0.658\\
ROUGE-2-R     & 0.937 & 0.952 & 0.841 & 0.746 & 0.820 & 0.654\\
ROUGE-3-P     & 0.828 & 0.940 & 0.800  & 0.610 & 0.839 & 0.656\\
ROUGE-L-F     & 0.865 & 0.604 & 0.461 &0.649 & 0.414 & 0.294\\
ROUGE-L-P     & 0.801 & 0.546 & 0.406 & 0.573 & 0.360 & 0.255\\
ROUGE-L-R     & 0.875 & 0.622 & 0.474 &0.663 & 0.414 & 0.298\\
ROUGE-W-1.2-F & 0.882 & 0.654 & 0.512 & 0.651 & 0.462 & 0.341\\
ROUGE-W-1.2-P & 0.798 & 0.514 & 0.393  & 0.558 & 0.337 & 0.237\\
ROUGE-W-1.2-R & 0.889 & 0.671 & 0.529 & 0.659 & 0.469 & 0.340\\
ROUGE-SU4-F   & 0.934 & 0.940 & 0.818  & 0.747 & 0.808 & 0.639\\
ROUGE-SU4-P   & 0.893 & 0.910 & 0.761  & 0.702 & 0.804 & 0.638\\
ROUGE-SU4-R   & 0.942 & 0.924 & 0.787 & 0.756 & 0.789 & 0.619\\

\hline
& \multicolumn{6}{c|}{AQUAINT-2 index ($\mathcal{I}=179,520$)} \\ \hline 
SERA-5     & 0.904 & 0.818 & 0.656 & 0.814 & 0.664 & 0.502 \\
SERA-10    & 0.881 & 0.817 & 0.651 & 0.813 & 0.675 & 0.513 \\
SERA-KW-5  & 0.900 & 0.816 & 0.654 & 0.807 & 0.665 & 0.503 \\
SERA-KW-10 & 0.880 & 0.810 & 0.646 &  0.807 & 0.670 & 0.511 \\
SERA-DIS-5 & 0.942 & 0.829 & 0.666 &  0.811 & 0.660 & 0.501 \\
SERA-DIS-NP-5 & 0.945 & 0.831 & 0.670 &  0.809 & 0.687 & 0.529 \\
SERA-DIS-KW-5 & 0.941 & 0.822 & 0.659 &  0.808 & 0.653 & 0.496 \\
SERA-DIS-KW-10 & 0.939 & 0.826 & 0.658 &  0.802 & 0.669 & 0.515 \\

\hline 
GeSERA-10 & 0.874 & 0.813 & 0.652 & 0.814 & 0.686 & 0.517 \\
GeSERA-DIS-10  & 0.940 & 0.818 & 0.657 & 0.818 & 0.673 & 0.514  \\ 
\hline
& \multicolumn{6}{c|}{Wikipedia index ($\mathcal{I}=10,000$)} \\ \hline 
SERA-5      & 0.942     & 0.870     & 0.717     & 0.843  & 0.768    & 0.592   \\
SERA-NP-5   & 0.926     & 0.863     & 0.704     & 0.831  & 0.749   & 0.573   \\
SERA-NP-10  & 0.936     & 0.863     & 0.709     &0.835   & 0.759   & 0.592  \\
SERA-KW-5   & 0.939     & 0.863     & 0.701     & 0.834  & 0.761   & 0.588   \\
SERA-DIS-5  & 0.952     & 0.877     & 0.729       & 0.809  & 0.778 & 0.602   \\
SERA-DIS-NP-5  &0.945     & 0.842     & 0.684       & 0.811  & 0.733 & 0.563  \\
SERA-DIS-NP-10 & 0.945     & 0.845     & 0.688       & 0.785  & 0.746 & 0.579  \\
SERA-DIS-KW-5  & 0.949     & 0.868     & 0.713       & 0.801  & 0.773 & 0.596  \\

\hline 
GeSERA-5   & 0.926  & 0.854  & 0.701 & 0.838   & 0.737   & 0.570 \\

\hline 
\end{tabular}
}
\caption{Correlations on TAC2008 and TAC2009.}
\label{tab:results_TAC_supp}
\end{table}

\end{document}